\documentclass[twoside,11pt]{article}
\PassOptionsToPackage{hyperfootnotes=false}{hyperref}
\usepackage[preprint]{jmlr2e}
\usepackage[T1]{fontenc}
\usepackage{lmodern}
\usepackage{amsmath}
\usepackage{booktabs}
\usepackage{array}
\usepackage{tabularx}
\usepackage{multirow}
\usepackage{microtype}
\usepackage{enumitem}
\usepackage[section]{placeins}
\usepackage{needspace}
\setlist{topsep=4pt,itemsep=3pt,parsep=0pt}
\newcommand{\BAcc}{\operatorname{BAcc}}

\newcommand{\argmax}{\operatorname*{arg\,max}}
\newcommand{\Top}{\operatorname{Top}}
\newcommand{\PSOIID}{\textsc{PSO-IID}}
\newcommand{\PSODA}{\textsc{PSO-DA50}}
\newcommand{\RSIID}{\textsc{RS-IID}}
\newcommand{\RSDA}{\textsc{RS-DA50}}
\newcolumntype{Y}{>{\raggedright\arraybackslash}X}
\newcolumntype{L}[1]{>{\raggedright\arraybackslash}p{#1}}
\ShortHeadings{Source-Domain Validation and Hyperparameter Selection}{Zaveri}
\firstpageno{1}
\hypersetup{pdftitle={When Source-Domain Validation Fails to Transfer: A Staged Study of Hyperparameter Selection},pdfauthor={Veet Zaveri},pdfsubject={Author-prepared research preprint}}

\begin{document}
\title{When Source-Domain Validation Fails to Transfer:\\
A Staged Study of Hyperparameter Selection\thanks{An abbreviated version of this work has been accepted to the AAAI-27 Student Abstract and Poster Program. This preprint combines the preliminary factorial study with the subsequent staged selection study and expands the methods, analysis, and discussion.}}
\author{\name Veet Zaveri \email vzaveri1@jhu.edu \\
\addr Johns Hopkins University\\Baltimore, MD, USA}
\maketitle

\begin{abstract}
Hyperparameters for an unseen domain are often selected using held-out examples from the available source domains. Withholding entire source domains during validation seems to better reflect deployment shift, but does this produce better selections on independent targets? We investigate this question through two complementary studies. A preliminary factorial experiment crosses particle swarm optimization and random search with pooled source validation and a leave-one-source-domain-out mean--minimum fitness on PACS and Multi-Domain Sentiment. The domain-aware fitness does not provide a consistent target advantage. We then separate selector development from confirmation using a shared pool of 64 configurations per task across four image benchmarks. Complete target-performance landscapes on VLCS and Office-Home inform one rule: retain the top 16 configurations by pooled source validation, then select the largest leave-one-source-domain-out mean-minus-standard-deviation score. The rule is frozen before evaluation on TerraIncognita and SVIRO. Across their 14 target domains, its equal-dataset-weighted balanced-accuracy difference from pooled validation is $+0.097$ percentage points (95\% bootstrap interval $[-0.685,+0.960]$), with five wins and nine losses. Exploratory analyses reveal substantial variation across datasets, target domains, and training seeds. These results do not establish equivalence or a general failure of domain-aware validation. They show that the evaluated source-domain criteria did not establish reliable target gains under the tested protocols, and motivate assessing selection rules on datasets independent of those used to develop them.
\end{abstract}

\begin{keywords}
domain generalization, hyperparameter selection, distribution shift, model selection, negative results, particle swarm optimization
\end{keywords}

\section{Introduction}
\label{sec:introduction}

A model can be selected without using any examples from its deployment domain and still be selected by an unreliable criterion. Consider a classifier trained on several visual styles or product categories. Pooled source validation asks which hyperparameters perform best on held-out examples from those familiar domains. Leave-one-source-domain-out (LODO) validation instead trains on a subset of the sources and evaluates on a source domain excluded from that fit. This resembles the intended deployment problem, but the observed source-to-source shifts need not represent the shift to a new target.

The resulting question concerns the \emph{selection rule}: the procedure that maps source-validation measurements to a hyperparameter configuration. Domain generalization studies prediction on domains unavailable during training \citep{muandet2013domain,wang2023survey}. Model selection is an established part of this problem; DomainBed explicitly compares source-based and oracle selection procedures and shows why the selection protocol must accompany the learning algorithm \citep{gulrajani2021lost}. Our focus is a narrower empirical question: \emph{does a rule that uses differences among observed source domains improve hyperparameter selection on datasets that did not inform the rule's design?}

Two kinds of separation matter. Within an outer-domain task, candidate fitting and source-validation scoring use only source examples and labels; fixed benchmark-preparation steps are described in Appendix~\ref{app:later-methods}. Across experiments, independent confirmation prevents target outcomes used to develop a selector from also serving as its evidence of transfer. A rule can satisfy the first requirement while having been designed using outcomes from the same benchmark on which it is evaluated. We therefore distinguish configuration selection from selection-rule development throughout the paper.

The investigation began with a factorial study of particle swarm optimization (PSO) and random search using either pooled source validation or a domain-aware mean--minimum fitness. Neither the original criterion nor an exploratory reweighting provided a consistent gain on PACS and Multi-Domain Sentiment. This motivated a second study with a common, fixed candidate pool. Complete target-performance landscapes on VLCS and Office-Home were used to develop a rule that retains a pooled-validation shortlist before applying a smoother LODO robustness score. We froze this rule before any source fitting or target evaluation on TerraIncognita and SVIRO. Further alternatives and diagnostics were examined only after this confirmation.

The paper addresses three research questions:
\begin{enumerate}[label=\textbf{RQ\arabic*.},leftmargin=3.2em]
\item Does a source-domain-aware fitness improve unseen-target selection under a fixed candidate budget, and does its effect depend on the search optimizer?
\item Does one selection rule developed on complete benchmark landscapes improve over pooled validation on previously untouched datasets?
\item After confirmation, what do proxy rankings, domain-specific outcomes, and training-seed variation reveal about the limits of the observed selections?
\end{enumerate}

The contribution is an empirical account of these questions with an explicit separation of evidence stages. The preliminary factorial design distinguishes optimizer behavior from the choice of objective; the later study holds the learner and candidate pool fixed while testing a frozen selector; and the exploratory analyses expose variation concealed by average effects. We do not propose LODO validation as a new principle or claim a universally superior selector. The main finding is that plausible source-domain criteria, even when optimized as intended, did not establish reliable target gains in these experiments.

\section{Related Work}
\label{sec:related}

\paragraph{Domain generalization and validation.}
Domain-generalization methods include invariant feature learning \citep{muandet2013domain}, adversarial representations \citep{li2018adversarial}, meta-regularization \citep{balaji2018metareg}, group distributionally robust training \citep{sagawa2020group}, risk extrapolation \citep{krueger2021rex}, and flat-minimum methods \citep{cha2021swad}. These approaches change model training. Here, the supervised learner is held fixed within each comparison, and the experimental intervention is how configurations are searched for or selected. Applying a robust score to validation outcomes is not the same procedure as robustly optimizing the training loss.

The closest methodological precedent is DomainBed \citep{gulrajani2021lost}, which studies pooled source validation, leave-one-domain-out validation, and target-domain oracle selection. Our work complements that comparison by following the development and independent confirmation of a particular source-based selector. It does not replicate DomainBed's training regime or establish a new general result about all LODO procedures. \citet{lu2023optimization} also study model selection under domain shift, constructing validation data through a Mixup-guided approach. We instead compare scoring and reranking rules on observed source domains; that method is not evaluated here.

\paragraph{Selection overfitting.}
Optimizing a validation criterion can exploit its sampling or training variability as well as its useful signal. \citet{cawley2010overfitting} analyze overfitting in model selection and the resulting bias in performance evaluation. This motivates our distinction between developing a selector and evaluating it independently. It also motivates treating retrospective rule comparisons as exploratory. The present experiments do not isolate validation overfitting as the cause of a particular target failure.

\paragraph{Hyperparameter search.}
PSO is a population-based search method whose particles combine personal and neighborhood information \citep{kennedy1995pso,clerc2002particle,kennedy2006neighborhood}. Prior applications establish its use in neural-network hyperparameter search \citep{ribalta2017pso,ye2017pso}. Random search supplies a simple comparison with the same candidate budget \citep{bergstra2012random}. Resource-allocation methods such as Hyperband and BOHB add another experimental factor \citep{li2018hyperband,falkner2018bohb}; we instead fix the number of optimizer updates per fit. The later image study removes adaptive candidate generation by comparing selectors on a common pool generated using a scrambled Sobol sequence \citep{sobol1967distribution}.

\section{Problem Formulation and Evidence Stages}
\label{sec:formulation}

\subsection{Outer targets, source validation, and the final learner}

Let a benchmark consist of domains $\mathcal D=\{D_1,\ldots,D_K\}$. In outer task $q$, domain $D_q$ is the target and $\mathcal S_q=\mathcal D\setminus\{D_q\}$ contains the sources. Each source is partitioned into training and validation examples. A configuration specifies hyperparameters; a \emph{selector} uses source-only scores to choose one configuration. The selected model is then trained afresh on all source examples and evaluated on the outer target.

For a labeled evaluation set containing $C$ classes, balanced accuracy is mean class recall:
\begin{equation}
\BAcc(M,E)=\frac{1}{C}\sum_{c=1}^{C}
\frac{\operatorname{TP}_c}{\operatorname{TP}_c+\operatorname{FN}_c}.
\label{eq:bacc}
\end{equation}
Here, $\operatorname{TP}_c$ and $\operatorname{FN}_c$ are true-positive and false-negative counts for class $c$. We express accuracies as percentages and differences as \emph{percentage points} (pp). A difference of $0.1$ pp is an absolute accuracy difference of $0.001$ on the $[0,1]$ scale.

For configuration $c$ and training seed $s$, write $I_s(c)$ for balanced accuracy on pooled source-validation examples after fitting on the pooled source-training data. The term ``IID'' denotes this within-source-mixture validation procedure; it does not assert that the unseen target follows the source distribution. Write $A_{e,s}(c)$ for balanced accuracy on source $e$'s validation examples after training an inner model on the other source domains. The outer target is excluded from both procedures. Pooled balanced accuracy weights classes equally but does not automatically give each source domain equal weight; LODO aggregation explicitly operates on separate domain scores.

\subsection{A chronological separation of evidence}

Table~\ref{tab:chronology} identifies which observations could influence each decision. In the development stage, target outcomes are deliberately inspected to design the next selector. Those datasets therefore provide development evidence. In the confirmation stage, the rule is fixed before any source fitting or target-performance evaluation on the two new datasets. ``Frozen'' refers to this documented ordering, not to a claim of public preregistration.

\begin{table}[htbp]
\centering\small
\caption{Experimental chronology. One frozen rule is tested in Stage III. Stage IV compares IID with nine alternatives, including the previously tested top-16 criterion.}
\label{tab:chronology}
\begin{tabularx}{\textwidth}{@{}L{0.15\textwidth}L{0.29\textwidth}Y@{}}
\toprule
Stage & Data and decision & Evidence status \\
\midrule
I: Preliminary & PACS and sentiment; PSO/RS $\times$ IID/DA50, followed by DA75 & Planned factorial comparisons; subsequent weighting analysis is exploratory. \\
\addlinespace
II: Development & VLCS and Office-Home; eight tasks with all 64 target outcomes & Target outcomes inform one top-16 mean-minus-SD selector. \\
\addlinespace
III: Confirmation & TerraIncognita and SVIRO; 14 tasks & Frozen selector versus IID on previously untouched datasets. \\
\addlinespace
IV: Diagnostics & All four later image datasets; 22 tasks & Retrospective proxy analysis and ten-rule exploratory comparison after confirmation outcomes are known. \\
\bottomrule
\end{tabularx}
\end{table}

The two studies are complementary but are not pooled. They differ in datasets, model families, candidate generation, and units of replication. The preliminary experiment has ten complete search repetitions per dataset. The later study uses one fixed candidate pool per task and repeated training seeds for its source scores and evaluated selections. Multiple training seeds do not create additional independent target domains.

\section{Preliminary Factorial Study}
\label{sec:preliminary}

\subsection{Design and the original domain-aware fitness}

The first study crosses optimizer $\{\mathrm{PSO},\mathrm{RS}\}$ with fitness $\{\mathrm{IID},\mathrm{DA50}\}$. PACS contains four visual domains \citep{li2017pacs}; Multi-Domain Sentiment contains four product-review domains \citep{blitzer2007biographies}. For each outer target and each of ten complete search repetitions, each arm evaluates 100 candidate configurations. Source splits and random seeds are paired where applicable, including the first ten candidate configurations. The subsequent adaptive PSO trajectories can differ by objective. Thus these arms share a candidate \emph{budget}, not the same complete candidate pool.

Let $A_e(\theta)$ denote inner held-out-source balanced accuracy for hyperparameters $\theta$, suppressing the outer-task and training-seed indices. The original domain-aware fitness is
\begin{equation}
J_{\mathrm{DA}}(\theta;\lambda)
=(1-\lambda)\frac{1}{m}\sum_{e\in\mathcal S_q}A_e(\theta)
+\lambda\min_{e\in\mathcal S_q}A_e(\theta),
\qquad m=|\mathcal S_q|=3.
\label{eq:da}
\end{equation}
DA50 fixes $\lambda=0.5$, giving equal coefficients to the mean and minimum terms. The minimum also contributes through the mean. The later exploratory DA75 condition sets $\lambda=0.25$: its name refers to a \emph{75\% mean / 25\% minimum} mixture. Neither criterion is the mean-minus-SD score used in the later frozen rule.

PACS uses ImageNet-pretrained ResNet-18 features through layer 3, with layer 4 and a classification head trained for each fit. Sentiment uses a TF--IDF multilayer perceptron. Source partitions are fixed and class-stratified at 80/20. Candidate and final fits each receive 200 optimizer updates. Selected configurations are trained afresh on all source examples. Appendix~\ref{app:preliminary-methods} specifies the representations, search spaces, PSO settings, and preprocessing safeguards.

The planned primary contrast (H1) is DA50 minus IID under PSO. Secondary contrasts examine the average fitness effect over both optimizers (H2a) and its optimizer interaction (H2b); a directional consistency criterion asks whether H1 is positive on both datasets (H3). Each inferential contrast uses ten paired repetition-level means, each averaging the four outer targets equally. Exact two-sided sign-flip tests operate on those ten differences. The original hierarchical bootstrap resamples repetitions and then target domains within each selected repetition; the two secondary tests are Holm-adjusted within dataset. Appendix~\ref{app:preliminary-results} gives all contrasts and their uncertainty.

\subsection{Target performance and optimization behavior}

Table~\ref{tab:preliminary-means} shows the original factorial outcomes. Under PSO, DA50 changed target balanced accuracy by $-0.121$ pp on PACS (95\% interval $[-1.025,+0.828]$, $p=0.7266$) and $-0.344$ pp on sentiment ($[-0.651,-0.045]$, $p=0.0059$). The effect was negative in nine of ten sentiment repetitions; PACS had five wins and five losses. The sentiment result is a small negative effect under this protocol. The PACS result provides no demonstrated advantage and leaves both harm and benefit compatible with its interval.

\begin{table}[htbp]
\centering\small
\caption{Preliminary factorial study: target balanced accuracy (\%). Parentheses are SDs across ten repetition-level means, each averaging four targets.}
\label{tab:preliminary-means}
\begin{tabular}{@{}lrr@{}}
\toprule
Condition & PACS & Multi-Domain Sentiment \\
\midrule
\PSOIID & $73.518\;(2.105)$ & $83.073\;(0.390)$ \\
\PSODA & $73.397\;(1.714)$ & $82.729\;(0.320)$ \\
\RSIID & $73.045\;(1.755)$ & $82.874\;(0.338)$ \\
\RSDA & $72.540\;(2.248)$ & $82.475\;(0.343)$ \\
\bottomrule
\end{tabular}
\end{table}

The mean DA50 effect across optimizers was also negative on sentiment ($-0.371$ pp; Holm-adjusted $p=0.0039$). Neither dataset showed a supported positive optimizer-by-fitness interaction. Although PSO achieved higher selected source fitness than random search under both objectives on both datasets, these source-score gains did not establish better target selection by DA50. The comparisons show that the source criterion could be improved within the search budget; they do not prove that its true optimum was found or isolate the cause of the source--target mismatch.

After the original results were examined, the DA75 sensitivity analysis produced a positive PACS point estimate relative to PSO-IID ($+0.310$ pp; interval $[-0.695,+1.431]$) but a negative sentiment estimate ($-0.408$ pp; $[-0.735,-0.103]$). The positive PACS mean was concentrated primarily in the sketch target and occurred despite only four of ten repetition-level wins. It did not identify a reliable replacement weighting.

The preliminary study also exposes a cost trade-off: every DA candidate needs three inner fits, whereas IID needs one. Equal candidate budgets therefore give DA three times as many candidate fits. The absence of a consistent target gain cannot be attributed to DA having a smaller number of candidate fits, but an equal-compute comparison remains a separate question.

\section{Fixed-Pool Development and Frozen Confirmation}
\label{sec:frozen}

\subsection{Benchmarks and common candidate protocol}

The second study considers VLCS \citep{fang2013}, Office-Home \citep{venkateswara2017}, TerraIncognita \citep{beery2018}, and SVIRO \citep{dacruz2020}. The evaluated benchmarks contain four, four, four, and ten domains, respectively, yielding 22 outer tasks. VLCS and Office-Home provide eight development tasks; TerraIncognita and SVIRO provide 14 confirmation tasks.

For each task, the compared selectors have access to the same pool of 64 scrambled-Sobol configurations, generated with scramble seed 20260920. The six dimensions are head learning rate, layer-4/head learning-rate ratio, weight decay, dropout, label smoothing, and AdamW $\beta_1$; all four datasets use the image bounds in Table~\ref{tab:search-spaces}. A shared ImageNet-pretrained ResNet-18 setup caches activations through layer 3 and trains layer 4 plus a task-specific head \citep{he2016deep}. Images are resized and center-cropped. The fixed source split is class-stratified 80/20, minibatches are class-balanced, and every candidate and final fit receives 200 optimizer updates. Appendix~\ref{app:later-methods} provides the implementation and seed settings.

Complete target-performance landscapes were obtained for all 64 configurations on each VLCS and Office-Home task, using two source-training seeds for source scores and two final-training seeds per configuration for target outcomes. These landscapes informed rule development. On the confirmation datasets, source selection uses three source-training seeds, and target evaluation initially concerned the IID and frozen selections, each with five matched final-training seeds. The same selected configuration is evaluated only once when rules agree. Consequently, the confirmation datasets do not supply full 64-configuration target landscapes.

\subsection{One frozen coarse-to-fine rule}
\label{sec:frozen-definition}

The rule retains the pooled-validation signal before introducing domain robustness. First, pooled source validation screens the 64 candidates to 16. Second, the LODO score reranks those 16. Let $m=|\mathcal S_q|$, with $m=3$ for TerraIncognita and $m=9$ for SVIRO. For source-training seeds $s=1,2,3$, define
\begin{align}
I(c)&=\frac{1}{3}\sum_{s=1}^{3}I_s(c),
&\bar A_s(c)&=\frac{1}{m}\sum_{e\in\mathcal S_q}A_{e,s}(c),\label{eq:source-scores}\\
R(c)&=\frac{1}{3}\sum_{s=1}^{3}
\left[\bar A_s(c)-\sqrt{\frac{1}{m}\sum_{e\in\mathcal S_q}
\bigl(A_{e,s}(c)-\bar A_s(c)\bigr)^2}\right].
\label{eq:robustness}
\end{align}
The domain SD uses population normalization, with denominator $m$. Domain variability is penalized within each seed before averaging the score over seeds. Let $\Top_{16}(\mathcal C;I)$ return the 16 configurations in $\mathcal C$ with the highest $I$ scores. The shortlist and selections are
\begin{equation}
\mathcal C_{16}=\Top_{16}(\mathcal C;I),\qquad
\hat c_{\mathrm{IID}}\in\argmax_{c\in\mathcal C}I(c),\qquad
\hat c_{\mathrm{F}}\in\argmax_{c\in\mathcal C_{16}}R(c).
\label{eq:selection}
\end{equation}
Score ties in both frozen rankings favor the candidate closest to the center of the normalized six-dimensional search space, then the lower configuration index.

The shortlist size and mean-minus-SD criterion were specified before any source fitting or target evaluation on TerraIncognita and SVIRO. Those datasets therefore test this one frozen rule, rather than a favorable rule selected retrospectively from several alternatives. The workflow is:
\begin{enumerate}[leftmargin=1.7em]
\item Exclude the outer target and construct source-training and validation partitions.
\item Score the shared candidate pool by pooled source validation, averaging over three source-training seeds.
\item Retain the top 16 configurations and choose the largest seed-averaged LODO mean-minus-SD score.
\item Fit the IID and frozen selections afresh on all source examples using five matched final-training seeds, then evaluate the target.
\end{enumerate}

The IID winner is itself in the shortlist. Maximizing $R$ therefore guarantees $R(\hat c_{\mathrm{F}})\geq R(\hat c_{\mathrm{IID}})$. A better source robustness score is an intended property of the selection procedure, not evidence by itself that target robustness improves. Target outcomes supply the separate empirical test.

\subsection{Primary estimand and uncertainty}

Let $B_{dt,s}(\hat c)$ be final target balanced accuracy on the $[0,1]$ scale for target $t$ of dataset $d$ using final-training seed $s$. The seed-averaged effect, in pp, is
\begin{equation}
\Delta_{dt}=100\left[\frac{1}{5}\sum_{s=1}^{5}
\bigl(B_{dt,s}(\hat c_{\mathrm{F}})-B_{dt,s}(\hat c_{\mathrm{IID}})\bigr)\right].
\label{eq:target-difference}
\end{equation}
The primary contrast gives the two confirmation datasets equal weight:
\begin{equation}
\widehat\Delta=\frac{1}{2}\sum_{d\in\{\mathrm{TerraIncognita},\mathrm{SVIRO}\}}
\frac{1}{T_d}\sum_{t=1}^{T_d}\Delta_{dt},
\qquad T_{\mathrm{TerraIncognita}}=4,\quad T_{\mathrm{SVIRO}}=10.
\label{eq:primary}
\end{equation}
SVIRO consequently does not receive more weight merely because it contains more evaluated domains. An unweighted mean over all 14 targets would estimate a different quantity.

The reported 95\% percentile bootstrap interval uses 20,000 resamples and seed 20260924, resampling target domains within each dataset while retaining equal dataset weights. An exact two-sided sign-flip analysis enumerates all $2^{14}=16{,}384$ domain-level sign assignments and recomputes the primary contrast. Five final-training seeds characterize stochastic training of each selected configuration; they are not 70 independent target-domain observations. These analyses summarize variation among the evaluated domain tasks conditional on the two datasets. The tasks reuse source domains, and the interval should not be interpreted as precise uncertainty over a broad population of new datasets.

\subsection{Confirmation results}

\begin{figure}[htbp]
\centering
\includegraphics[width=\textwidth]{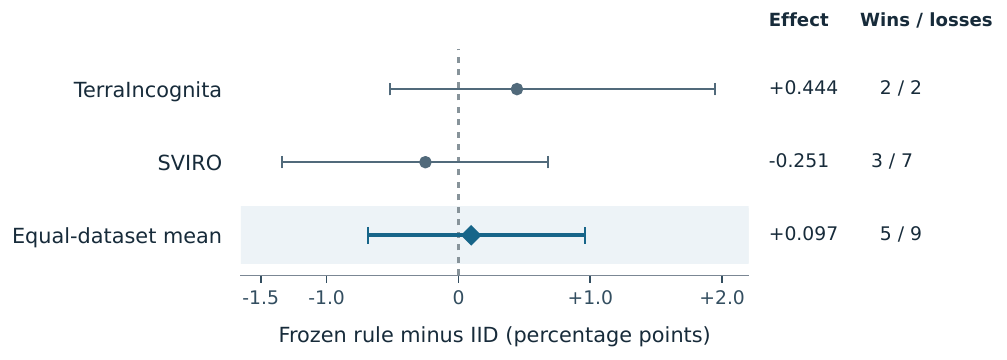}
\caption{Frozen rule minus pooled IID on previously untouched datasets. Bars show reported 95\% bootstrap intervals; dataset rows are descriptive, and the primary row weights the two datasets equally. Counts refer to domain wins and losses. Positive values favor the frozen rule.}
\label{fig:confirmation}
\end{figure}

The frozen rule achieved $+0.444$ pp on TerraIncognita (two wins, two losses) and $-0.251$ pp on SVIRO (three wins, seven losses). The primary equal-dataset estimate was $+0.097$ pp, with a 95\% bootstrap interval of $[-0.685,+0.960]$ and sign-flip $p=0.837$ (Figure~\ref{fig:confirmation}). The rule was not confirmed as an improvement. The interval includes both harm and benefit; it does not establish equivalence.

The small average conceals domain-specific outcomes. The frozen choice gained $+2.725$ pp on TerraIncognita/Location38 but lost $-4.272$ pp on SVIRO/X5; X5 lost in four of five matched final-training seeds. Table~\ref{tab:frozen-domains} reports all 14 target outcomes. The source-side audit verified the intended argmax selections and that target fits began after source selection was complete. All 14 chosen configurations strictly improved the aggregate LODO robustness score relative to IID, yet only five improved mean target performance. This establishes a mismatch between source-score preference and target benefit in the observed comparisons, without identifying its causal mechanism.

A post-hoc grouping by the selected configuration's IID rank gives five wins and three losses for ranks 2--8, compared with no wins and six losses for ranks 9--16. This suggests that departures further down the IID ranking deserve scrutiny. It does not estimate the performance of a top-eight replacement selector: that selector would make its own choices, and its design would now be informed by the confirmation outcomes.

\section{Exploratory Analyses After Confirmation}
\label{sec:exploratory}

The following analyses use datasets whose target outcomes had already been examined. They describe the behavior of the measured pools and selections, and do not constitute another untouched confirmation.

\subsection{Global proxy ranking versus the selected configuration}

Full proxy--target ranking can be assessed only for the eight development tasks, where all 64 candidate target outcomes are available. Table~\ref{tab:correlation} reports mean Spearman correlations for IID, LODO mean, and LODO mean-minus-SD. Both LODO proxies correlated more strongly with target performance than pooled IID on Office-Home, but less strongly on VLCS. The aggregate therefore conceals dataset dependence.

\begin{table}[htbp]
\centering\small
\caption{Development-landscape diagnostics: mean Spearman source-proxy/target correlation across all 64 candidates. These are full-pool proxy diagnostics; the mean-minus-SD row does not include the frozen rule's IID shortlist.}
\label{tab:correlation}
\begin{tabular}{@{}lrrr@{}}
\toprule
Source proxy & VLCS & Office-Home & Equal-dataset mean \\
\midrule
Pooled IID & 0.806 & 0.941 & 0.874 \\
LODO mean & 0.668 & 0.980 & 0.824 \\
LODO mean minus SD & 0.524 & 0.969 & 0.747 \\
\bottomrule
\end{tabular}
\end{table}

A high correlation over the entire pool does not guarantee the best choice near the top. Define measured target regret as the balanced-accuracy gap between the best observed candidate in the evaluated pool and the candidate selected by a source proxy. Across the eight development tasks, mean regret was $0.943$ pp for pooled IID, $2.063$ pp for LODO mean, and $1.161$ pp for unshortlisted LODO mean-minus-SD. This is a comparison against a finite, noisy measured pool, not an estimate of regret relative to the unknown optimal hyperparameters. The best-minus-median candidate gap ranged from $2.769$ to $9.533$ pp across these tasks: the pools were not flat, although the remaining measured gain over IID was much smaller on average.

These results are descriptive because the landscapes informed rule development. On TerraIncognita and SVIRO, measuring correlations only among configurations selected for target evaluation would condition on previous selection. We therefore do not present those partial observations as full-landscape correlations.

\subsection{One frozen rule and a later ten-rule comparison}
\label{sec:exploratory-family}

The later exploratory analysis, called Stage C in the experimental records, compares \emph{ten rules in total}: the pooled-IID baseline and nine alternatives. Every alternative uses the seed-averaged LODO mean-minus-population-SD score. Four rerank IID top-2, top-4, top-8, or top-16 shortlists; two restrict candidates to the top 16 and to source-IID regret of at most 0.5 or 1.0 pp; and three penalize IID regret, optionally with an additional source-seed variability penalty. Appendix~\ref{app:exploratory-results} defines all rules and reports all ten outcomes. The family includes the previously tested top-16 criterion; these are not nine additional independent confirmations.

Stage C reuses two-seed source summaries on VLCS and Office-Home and three-seed source summaries on TerraIncognita and SVIRO. New configurations selected on TerraIncognita and SVIRO received five-seed target evaluations, with exact existing outcomes reused. VLCS and Office-Home reused their earlier full landscapes with two final-training seeds per configuration. Equal-dataset averages therefore combine different amounts of training-seed measurement, while giving each benchmark equal weight.

The top-16 variant had an equal-dataset mean effect of $+0.086$ pp across all four datasets; the IID-regret-penalized variant with penalty coefficient 1 had $+0.062$ pp. Their descriptive intervals included zero, and both had more task losses than wins. The Stage C top-16 variant reproduces all 14 frozen confirmation selections and target effects. Its $+0.086$ pp four-dataset mean additionally includes VLCS and Office-Home; it is a separate exploratory aggregate, not a change to the $+0.097$ pp two-dataset confirmation estimate. Formal tie-breaking conventions are specified in Appendix~\ref{app:exploratory-results}.

No non-IID rule passed the stated advancement criteria: at least $+0.1$ pp equal-dataset improvement, nonnegative effects on at least three of four datasets, more task wins than losses, and a positive leave-one-dataset-out rule-selection result with more dataset wins than losses. These are study-specific progression criteria, not significance tests or a guarantee of generalization. The recorded decision was to retain IID rather than advance another rule to confirmation.

\subsection{Exploratory rule transfer across datasets}

\begin{figure}[htbp]
\centering
\includegraphics[width=0.92\textwidth]{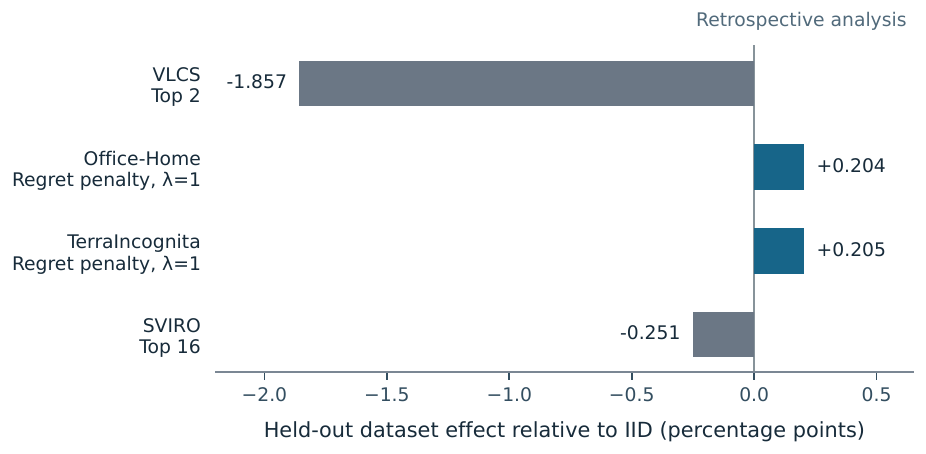}
\caption{Exploratory leave-one-dataset-out assessment of rule selection. Each bar is the effect on one dataset of a rule selected using the other three. All datasets had already been examined during the broader investigation; this is a retrospective transfer diagnostic. The mean is $-0.425$ pp, with two gains and two losses.}
\label{fig:cross-dataset}
\end{figure}

To examine sensitivity to the development benchmarks, an exploratory four-fold analysis selected a rule using three datasets and scored it on the fourth. An alternative was eligible when its mean effect was nonnegative on at least two of the three training datasets and its equal-dataset mean was nonnegative; IID was always eligible. Selection maximized the training-fold mean, with the tie-breakers specified in Appendix~\ref{app:exploratory-results}. The held-out effects were $-1.857$ pp on VLCS, $+0.204$ on Office-Home, $+0.205$ on TerraIncognita, and $-0.251$ on SVIRO, giving an equal-dataset mean of $-0.425$ pp. Figure~\ref{fig:cross-dataset} identifies each selected rule.

Holding out one dataset within this retrospective analysis reduces direct use of that fold's outcomes in its rule choice. It does not undo earlier exposure of the overall investigation to all four datasets. Four folds also provide little information about performance across a wider population of benchmarks.

\subsection{Training-seed sensitivity}

Across 13 evaluated configuration--target groups on TerraIncognita and 32 on SVIRO, the median within-configuration sample SD of target balanced accuracy was $2.253$ and $1.166$ pp, respectively. These are descriptive summaries of the evaluated selections, not uncertainty estimates for the full unobserved candidate pools.

An \emph{override} occurs when an exploratory rule chooses a different configuration from IID. On the confirmation datasets, matched-seed effect differences changed sign across the five seeds in 10 of 12 override tasks for the penalty-1 anchor and 11 of 14 for the exploratory top-16 variant. The selected configuration could outperform IID under one training seed and underperform it under another. This variability helps explain why a small mean effect warrants caution, but does not demonstrate that training noise alone caused the negative findings.

\section{Discussion}
\label{sec:discussion}

\subsection{What the evidence supports}

The studies provide a bounded negative result. Under the initial factorial protocol, the mean--minimum criterion did not establish better target selection and produced a small adverse sentiment effect. Under the later fixed-pool protocol, the frozen mean-minus-SD shortlist rule did not establish an average improvement on the two untouched datasets. Some individual targets benefited, and both the PACS and frozen-confirmation intervals remain compatible with positive average effects of limited size. The evidence therefore supports neither universal failure nor statistical equivalence.

The practical implication is that extra source-domain evaluation must be justified by measured target benefit. Its resemblance to deployment shift is not sufficient. For researchers designing selectors, the useful question is whether a rule improves selection \emph{after the development benchmarks change}. The frozen confirmation addresses that question for one rule and two datasets; the exploratory results show why repeatedly modifying the rule on the same opened benchmarks would not supply independent evidence.

\subsection{Plausible explanations, not established mechanisms}

Several mechanisms are compatible with the results. First, observed sources may not represent the target shifts that matter. Transfer among some product categories, camera locations, or visual styles may favor hyperparameters that are unhelpful on another. A source-domain aggregate can reward consistency across the available sources without identifying the relevant invariances for the next domain.

Second, LODO changes the fitting problem as well as the validation distribution. Each inner model uses one fewer source domain than the final model. In the four-domain benchmarks, this means two inner training domains versus three in final training; in SVIRO it means eight versus nine. The number and mixture of available training examples also change. Pooled validation more closely matches the final source mixture, even though its validation examples do not simulate an unseen domain. The experiments do not disentangle this training-set difference from the value of holding out a source.

Third, a minimum or variability penalty can be sensitive to noisy domain measurements. Averaging over source-training seeds reduces one source of variation but does not make the score an exact estimate of future-domain performance. Selecting a maximum among noisy candidates can amplify favorable noise \citep{cawley2010overfitting}. The target-seed diagnostics document instability; they do not quantify how much is attributable to noisy source selection, final training, or the domain mismatch itself.

Finally, global proxy quality and top-of-pool selection quality differ. A proxy can rank clearly poor configurations correctly while reversing the order of a few strong candidates. The development-landscape correlations and measured regret illustrate why an apparently high correlation is not, by itself, evidence of a useful selector.

\subsection{Cost and experimental scope}

The preliminary DA searches use three candidate fits per evaluated configuration, versus one for IID. In the later study, the number of LODO inner fits depends on the number of sources, and screening can limit which configurations need those fits. Shared candidates and a fixed per-fit update budget do not equalize total computation. We therefore report the original fit-count comparison, not a threefold wall-clock claim for every experiment. No equal-total-compute superiority comparison was performed.

The later study covers four image benchmarks with one pretrained backbone, one candidate-space design, and a short 200-update fitting budget. It does not test fully end-to-end image training, multiple modern backbones, or language-model representations. The initial sentiment experiment broadens the motivating evidence to text, but it uses a different learner and search procedure and cannot establish cross-modal transfer of the frozen image selector.

Only eight development tasks have complete target landscapes. The confirmation evaluates selected candidates, which answers the prespecified comparison but leaves their positions in the unmeasured target landscape unknown. The two confirmation datasets, overlapping source sets, and unequal numbers of final-training seeds across development and confirmation further limit population-level inference. None of the experiments was designed to establish equivalence within a prespecified margin.

\subsection{Implications for subsequent evaluation}

A useful next study would fix a new selector and its decision criteria before examining another dataset, compare methods at matched total candidate-fit or training-example budgets, and vary the backbone and training budget. Repeated source-seed estimates and complete target landscapes on development tasks could help distinguish unstable proxy rankings from stable but inappropriate rankings. Any new rule suggested by the present shortlist or weighting diagnostics would require a fresh confirmation. These are proposed extensions, not experiments reported in this paper.

\Needspace{14\baselineskip}
\section{Conclusion}
\label{sec:conclusion}

We investigated source-domain-aware hyperparameter selection through a preliminary factorial search study, fixed-pool selector development, one untouched-dataset confirmation, and subsequent exploratory diagnostics. The original mean--minimum fitness and the frozen IID-shortlist/mean-minus-SD rule did not establish reliable target gains under their respective protocols. Better observed source scores did not consistently identify better unseen-target configurations, and retrospective alternatives varied across datasets and training seeds. The central lesson is methodological: a selector's source-only implementation and its independent evaluation are separate requirements. Reporting both makes the scope of a negative or mixed result clearer and prevents development findings from being mistaken for evidence of transfer.

\FloatBarrier
\appendix

\section{Preliminary Study: Reproducible Protocol Details}
\label{app:preliminary-methods}

This appendix specifies the PACS and Multi-Domain Sentiment protocol used in the preliminary factorial study.

\subsection{Data, representations, and preprocessing}

PACS contains 9,991 images from seven classes: dog, elephant, giraffe, guitar, horse, house, and person. Domain sizes are 2,048 art-painting images, 2,344 cartoon images, 1,670 photographs, and 3,929 sketches \citep{li2017pacs}. Images are resized to a shorter side of 256 pixels, center-cropped to $224\times224$, and normalized using ImageNet statistics, without augmentation. ResNet-18's stem and layers 1--3 are frozen, with deterministic layer-3 activations of shape $256\times14\times14$ cached. Layer 4 starts from its pretrained weights and is trained alongside adaptive average pooling, dropout, and a newly initialized seven-class linear head.

Multi-Domain Sentiment has 8,000 labeled reviews across books, DVDs, electronics, and kitchen products, with 2,000 reviews and balanced positive/negative labels per domain \citep{blitzer2007biographies}. The supplied unigram/bigram counts are transformed using TF--IDF. The vocabulary is limited to 20,000 features with minimum document frequency 2; features are ordered by decreasing document frequency, with lexicographic tie-breaking. For $N$ allowed training documents,
\begin{equation}
\operatorname{IDF}(t)=\log\!\left(\frac{1+N}{1+\operatorname{df}(t)}\right)+1,
\end{equation}
followed by document-level $\ell_2$ normalization. Vocabulary and IDF are fitted only on the training data allowed for that fit. Each DA inner holdout therefore refits them using its two training domains. The multilayer perceptron has dimensions $20{,}000\rightarrow512\rightarrow128\rightarrow2$, GELU hidden activations, and separately tuned hidden-layer dropout.

Within each source domain, fixed class-stratified 80/20 partitions use manifest seed 271828. Candidate and final models minimize cross-entropy for exactly 200 optimizer updates, without early stopping, using the configuration's label smoothing and a constant nominal learning rate within each fit. AdamW uses $\beta_2=0.999$ and $\epsilon=10^{-8}$, with weight decay applied only to tensors of at least two dimensions \citep{kingma2015adam,loshchilov2019decoupled}. PACS uses class-balanced minibatches of 64 and evaluation batches of 256; sentiment uses 128 and 512. Deterministic algorithms are enabled and TensorFloat-32 is disabled. Final fits use the union of source-training and source-validation examples, with preprocessing refitted on those sources when required.

\subsection{Search spaces and PSO}

Both datasets use six-dimensional continuous search spaces (Table~\ref{tab:search-spaces}). Log-uniform parameters are searched in logarithmic coordinates; all coordinates are normalized to $[0,1]$ for PSO.

\begin{table}[htbp]
\centering\small
\caption{Search bounds. The image column is used for preliminary PACS and for all four later image benchmarks. A dash indicates a dimension absent from that six-dimensional space.}
\label{tab:search-spaces}
\begin{tabular}{@{}lrr@{}}
\toprule
Parameter & Images & Sentiment \\
\midrule
Head/overall learning rate (log) & $[10^{-5},10^{-2.5}]$ & $[10^{-5},10^{-2}]$ \\
Layer-4/head learning-rate ratio (log) & $[10^{-2},1]$ & --- \\
Weight decay (log) & $[10^{-6},10^{-2}]$ & $[10^{-7},10^{-2}]$ \\
First/only dropout & $[0,0.60]$ & $[0,0.70]$ \\
Second hidden-layer dropout & --- & $[0,0.70]$ \\
Label smoothing & $[0,0.20]$ & $[0,0.20]$ \\
AdamW $\beta_1$ & $[0.80,0.99]$ & $[0.80,0.99]$ \\
\bottomrule
\end{tabular}
\end{table}

The swarm has ten particles in a ring, with neighbors $\{i-1,i,i+1\}$ under circular indexing. Its updates are
\begin{align}
v_i^{(t+1)}&=\omega_t v_i^{(t)}+c_1r_1\odot(p_i-x_i^{(t)})
+c_2r_2\odot(n_i-x_i^{(t)}),\\
x_i^{(t+1)}&=x_i^{(t)}+v_i^{(t+1)},
\end{align}
where $p_i$ and $n_i$ are personal and neighborhood best positions and $r_1,r_2$ are independent coordinatewise $\operatorname{Uniform}(0,1)$ draws at each update. Both coefficients are 1.5. Initial velocities are uniform on $[-0.1,0.1]$, velocity bounds are $[-0.2,0.2]$, and boundaries are reflective. The initial swarm plus nine updates provides 100 evaluations. Inertia decreases linearly from 0.9 on the first update to 0.4 on the ninth. Fitness ties favor the candidate closest to the normalized search-space center, then the earlier evaluation index.

Within each target--repetition block, conditions share source manifests, repetition seeds, the first ten configurations, candidate-index training seeds, and final-training seeds where applicable. Random search adds 90 unique uniform samples after the common initial ten. The complete search seeds are 1103, 2207, 3301, 4409, 5519, 6619, 7727, 8837, 9941, and 11047. The fixed manifests do not change with the search repetition, so replication characterizes search and training variation conditional on the observed data.

\subsection{Workload and inference}

Across both planned factorial experiments, the workload is 32,000 candidate evaluations, 64,000 candidate fits, 320 final fits, and 12,864,000 optimizer updates. Each dataset's exploratory DA75 run adds 4,000 evaluations, 12,000 candidate fits, 40 final fits, and 2,408,000 updates. These counts describe work performed across all arms; they do not imply 32,000 distinct shared configurations.

Let $d_{o}$ be the DA50-minus-IID effect for optimizer $o$ in one dataset, with each repetition averaged equally over its four targets. The planned contrasts are $d_{\mathrm{PSO}}$ (H1), $(d_{\mathrm{PSO}}+d_{\mathrm{RS}})/2$ (H2a), and $d_{\mathrm{PSO}}-d_{\mathrm{RS}}$ (H2b). The directional hypotheses are positive effects; the reported tests are two-sided. H3 requires positive H1 point estimates on both datasets and is a consistency criterion rather than a pooled test.

Exact sign-flip tests enumerate all $2^{10}$ assignments of the ten paired repetition-level differences. The hierarchical bootstrap uses 10,000 samples and seed 314159, resampling repetitions and then domains within each selected repetition. H1 is unadjusted; H2a and H2b are Holm-adjusted together within each dataset. These uncertainty calculations follow the original analysis and do not make the overlapping outer-domain tasks independent population samples. Target-specific effects, worst-target performance, and across-target SD are descriptive. The latter two are computed within each repetition before averaging, so the mean worst-target score need not equal the minimum of target-specific mean scores.

\section{Preliminary Study: Additional Results}
\label{app:preliminary-results}

Tables~\ref{tab:planned-contrasts}--\ref{tab:source-fitness} provide the planned contrasts, secondary metrics, target-specific outcomes, and source-objective comparisons. The following sensitivity analysis reports the later exploratory DA75 run.

\begin{table}[htbp]
\centering\small
\caption{Complete planned preliminary contrasts, in pp. H2 $p$-values are Holm-adjusted within dataset. Intervals and tests retain the original repetition-based analysis.}
\label{tab:planned-contrasts}
\begin{tabular}{@{}llrrr@{}}
\toprule
Dataset & Contrast & Estimate & 95\% interval & $p$ \\
\midrule
PACS & H1: PSO DA50 $-$ IID & $-0.121$ & $[-1.025,+0.828]$ & 0.7266 \\
& H2a: mean DA50 $-$ IID & $-0.313$ & $[-1.583,+1.068]$ & 1.0000 \\
& H2b: optimizer interaction & $+0.385$ & $[-1.654,+2.341]$ & 1.0000 \\
\addlinespace
Sentiment & H1: PSO DA50 $-$ IID & $-0.344$ & $[-0.651,-0.045]$ & 0.0059 \\
& H2a: mean DA50 $-$ IID & $-0.371$ & $[-0.645,-0.114]$ & 0.0039 \\
& H2b: optimizer interaction & $+0.055$ & $[-0.315,+0.468]$ & 0.6641 \\
\bottomrule
\end{tabular}
\end{table}

\begin{table}[htbp]
\centering\small
\caption{Secondary target metrics (\%). Parentheses are SDs across ten repetition-level means. Balanced accuracy, the primary metric, appears in Table~\ref{tab:preliminary-means}.}
\label{tab:secondary-metrics}
\begin{tabular}{@{}llrr@{}}
\toprule
Dataset & Condition & Accuracy & Macro-$F_1$ \\
\midrule
PACS & \PSOIID & $72.529\;(1.867)$ & $72.313\;(1.928)$ \\
& \PSODA & $72.456\;(1.686)$ & $72.313\;(1.746)$ \\
& \RSIID & $71.981\;(1.311)$ & $71.822\;(1.596)$ \\
& \RSDA & $71.223\;(2.323)$ & $71.302\;(2.502)$ \\
\addlinespace
Sentiment & \PSOIID & $83.073\;(0.390)$ & $83.035\;(0.399)$ \\
& \PSODA & $82.729\;(0.320)$ & $82.689\;(0.329)$ \\
& \RSIID & $82.874\;(0.338)$ & $82.830\;(0.343)$ \\
& \RSDA & $82.475\;(0.343)$ & $82.432\;(0.352)$ \\
\bottomrule
\end{tabular}
\end{table}

\begin{table}[htbp]
\centering\small
\caption{Target-specific balanced accuracy (\%), averaged over ten repetitions. These domain-level summaries are descriptive.}
\label{tab:domain-means}
\begin{tabular}{@{}llrrrr@{}}
\toprule
Dataset & Target & \PSOIID & \PSODA & \RSIID & \RSDA \\
\midrule
PACS & Art painting & 77.258 & 77.029 & 76.075 & 77.175 \\
& Cartoon & 70.751 & 69.829 & 70.026 & 68.354 \\
& Photo & 93.209 & 93.686 & 93.786 & 93.561 \\
& Sketch & 52.856 & 53.045 & 52.294 & 51.069 \\
\addlinespace
Sentiment & Books & 79.965 & 80.030 & 79.860 & 79.495 \\
& DVD & 80.585 & 80.390 & 80.420 & 80.200 \\
& Electronics & 84.410 & 83.305 & 84.105 & 83.480 \\
& Kitchen & 87.330 & 87.190 & 87.110 & 86.725 \\
\bottomrule
\end{tabular}
\end{table}

\Needspace{8\baselineskip}
Under PSO, the largest sentiment loss was on electronics ($-1.105$ pp), where DA50 never exceeded IID across the ten repetitions. PACS worst-target balanced accuracy was $52.856\%$ for IID and $53.045\%$ for DA50; sentiment's mean worst-target score was $79.850\%$ for both. The mean within-repetition across-target SD changed from 16.810 to 17.055 pp on PACS and from 3.479 to 3.368 pp on sentiment. These outcomes do not indicate a consistent improvement in target robustness.

\begin{table}[htbp]
\centering\small
\caption{Mean selected source fitness (\%) over 40 target--repetition blocks per dataset. Compare optimizers \emph{within} each objective; IID and DA50 scores use different training and evaluation procedures.}
\label{tab:source-fitness}
\begin{tabular}{@{}llrr@{}}
\toprule
Dataset & Fitness & PSO & Random search \\
\midrule
PACS & IID & 94.420 & 94.162 \\
& DA50 & 65.574 & 65.081 \\
Sentiment & IID & 87.731 & 87.369 \\
& DA50 & 79.522 & 79.188 \\
\bottomrule
\end{tabular}
\end{table}

\FloatBarrier
\Needspace{10\baselineskip}
\subsection{Exploratory DA75 sensitivity analysis}

Only PSO-DA75 was rerun after the original results were examined. Targets, manifests, repetition seeds, model architectures, PSO settings, candidate budget, and training schedule remained fixed. Its source fitness is
\begin{equation}
J_{\mathrm{DA75}}(\theta)=0.75\bar A(\theta)+0.25\min_e A_e(\theta).
\end{equation}
Here, $\bar A(\theta)$ is the mean of the three inner held-out-source scores. Table~\ref{tab:da75} reports its outcomes. PACS exceeded IID in four of ten repetitions and DA50 in six. Sentiment trailed IID in eight repetitions. DA75's favorable PACS mean was driven primarily by sketch ($+1.735$ pp versus IID), while sentiment electronics remained unfavorable ($-1.365$ pp and negative in every repetition).

\begin{table}[htbp]
\centering\small
\caption{Exploratory PSO-DA75 analysis. Balanced accuracy and worst-target means are percentages; contrasts and intervals are pp. Reported $p$-values are exploratory and do not establish a new confirmed rule.}
\label{tab:da75}
\begin{tabular}{@{}llrrr@{}}
\toprule
Dataset & Summary or contrast & Value & 95\% interval & $p$ \\
\midrule
PACS & Balanced accuracy & 73.828 & --- & --- \\
& Worst-target mean & 54.590 & --- & --- \\
& DA75 $-$ IID & $+0.310$ & $[-0.695,+1.431]$ & 0.4941 \\
& DA75 $-$ DA50 & $+0.431$ & $[-0.398,+1.405]$ & 0.3750 \\
\addlinespace
Sentiment & Balanced accuracy & 82.665 & --- & --- \\
& Worst-target mean & 79.840 & --- & --- \\
& DA75 $-$ IID & $-0.408$ & $[-0.735,-0.103]$ & 0.0117 \\
& DA75 $-$ DA50 & $-0.064$ & $[-0.263,+0.128]$ & 0.4199 \\
\bottomrule
\end{tabular}
\end{table}

\FloatBarrier
\section{Later Study: Implementation and Confirmation Details}
\label{app:later-methods}

\subsection{Candidate pool and training configuration}

All four later image benchmarks use the six image bounds in Table~\ref{tab:search-spaces}. The 64-point scrambled Sobol design uses seed 20260920 and is regenerated identically in normalized coordinates for each task. For a coordinate $u\in[0,1]$ and bounds $[a,b]$, the mapping is $a+u(b-a)$ for a linear parameter and $10^{\log_{10}a+u(\log_{10}b-\log_{10}a)}$ for a logarithmic parameter. Thus the selectors compare identical candidate hyperparameters within a task, rather than separate random searches.

The archived implementation uses ImageNet-1K V1 ResNet-18 weights, with the stem and layers 1--3 frozen. Images are resized to a shorter side of 256 pixels, center-cropped to $224\times224$, and normalized with ImageNet statistics, without augmentation. Cached layer-3 activations have shape $256\times14\times14$ and are stored in float16; trainable-model computation uses float32. Layer 4 starts from pretrained weights and is trained with adaptive average pooling, the selected dropout, and a new classifier for the benchmark's classes.

Source train/validation partitions are class-stratified 80/20 within each domain using seed 271828. Exact duplicate groups are kept within a single partition; cross-domain or conflicting-label exact duplicate groups are excluded. Dataset preparation includes this fixed audit across domains and fixed pretrained feature extraction. After preparation, outer-target examples and labels are excluded from candidate training and source-validation scoring, and target-performance outcomes remain unavailable until the confirmation choices are fixed. Pooled IID fits use all source-training partitions and evaluate their pooled validation partitions. Each LODO fit trains on the training partitions of all remaining sources and evaluates the excluded source's validation partition. Final fits use all available source examples.

Each fit receives exactly 200 updates with cross-entropy and the selected label smoothing. AdamW uses the selected $\beta_1$, fixed $\beta_2=0.999$ and $\epsilon=10^{-8}$, with weight decay only on tensors of at least two dimensions. There is no early stopping or learning-rate schedule. The class-balanced training batch sizes are 64 for VLCS, 130 for Office-Home, 80 for TerraIncognita, and 70 for SVIRO; evaluation uses batches of 256. Deterministic algorithms are enabled and TensorFloat-32 is disabled. These settings define a short, partly frozen-backbone training protocol, not full end-to-end convergence.

\subsection{Seeds, fitting cost, and target summaries}

Table~\ref{tab:later-seeds} gives the experiment-level repetition seeds. Actual fit seeds are derived deterministically from these seeds and dataset, target, and fit identifiers. Source fits also include the candidate index and, where applicable, the inner fold. Final-fit seed derivation omits candidate identity, so different selected configurations receive matched final-training seeds. Stage C reuses the earlier source summaries and existing target outcomes, evaluating only distinct newly selected configurations when necessary.

\begin{table}[htbp]
\centering\small
\caption{Repetition seeds in the later study. Development uses two source-score and two final-training seeds; confirmation uses three and five. Stage C retains these dataset-specific seed sets.}
\label{tab:later-seeds}
\begin{tabularx}{\textwidth}{@{}L{0.27\textwidth}Y Y@{}}
\toprule
Dataset group & Source-score seeds & Final-training seeds \\
\midrule
VLCS, Office-Home & 13007, 23003 & 41011, 51001 \\
TerraIncognita, SVIRO & 13007, 23003, 33013 & 71011, 81001, 91009, 101003, 111017 \\
\bottomrule
\end{tabularx}
\end{table}

In frozen confirmation, each task requires $64\times3=192$ pooled-IID source fits. The additional shortlist evaluation requires $16\times3\times m$ LODO fits, where $m=3$ for TerraIncognita and $m=9$ for SVIRO. Across the 14 tasks this totals 2,688 IID fits plus 4,896 LODO fits, or 7,584 source fits. The final target evaluations are additional. These counts make the extra selection cost explicit without equating fit counts with wall-clock time.

Table~\ref{tab:frozen-domains} reports every frozen-confirmation target. The primary analysis averages the five matched-seed differences within a task, then averages tasks within each dataset, then weights the two datasets equally. It does not weight a domain by its image count or treat training seeds as additional domains. Sign counts use the direction of the mean effect, which can differ from the majority direction across seeds when effect magnitudes vary.

\begin{table}[htbp]
\centering\small
\caption{Complete frozen-confirmation outcomes. Accuracies are percentages averaged over five matched final-training seeds; effects are frozen minus IID in pp. The last column counts positive paired seed effects and is descriptive. Rounding can make displayed differences differ by 0.001 pp from subtraction of the displayed means.}
\label{tab:frozen-domains}
\begin{tabular}{@{}llrrrr@{}}
\toprule
Dataset & Target & IID (\%) & Frozen (\%) & Effect (pp) & Positive seeds \\
\midrule
TerraIncognita & Location 100 & 30.762 & 30.104 & $-0.658$ & 3/5 \\
& Location 38 & 20.433 & 23.158 & $+2.725$ & 5/5 \\
& Location 43 & 23.035 & 23.127 & $+0.091$ & 3/5 \\
& Location 46 & 22.031 & 21.647 & $-0.383$ & 2/5 \\
\addlinespace
SVIRO & aclass & 98.953 & 98.691 & $-0.262$ & 0/5 \\
& escape & 98.607 & 97.954 & $-0.654$ & 0/5 \\
& hilux & 96.887 & 97.039 & $+0.151$ & 4/5 \\
& i3 & 85.915 & 85.791 & $-0.125$ & 4/5 \\
& lexus & 98.198 & 98.131 & $-0.068$ & 4/5 \\
& tesla & 97.404 & 96.770 & $-0.634$ & 2/5 \\
& tiguan & 84.297 & 85.329 & $+1.032$ & 3/5 \\
& tucson & 95.231 & 97.784 & $+2.553$ & 4/5 \\
& x5 & 93.714 & 89.443 & $-4.272$ & 1/5 \\
& zoe & 95.483 & 95.254 & $-0.229$ & 2/5 \\
\bottomrule
\end{tabular}
\end{table}

\FloatBarrier
\section{Later Study: Complete Exploratory Rule Comparison}
\label{app:exploratory-results}

\subsection{Scores, eligibility, and tie-breaking}

The exploratory comparison contains one IID baseline and nine alternatives. Let $S=2$ for VLCS and Office-Home and $S=3$ for TerraIncognita and SVIRO. Write $r_s(c)=\bar A_s(c)-\sigma_s(c)$, where $\sigma_s$ is the population SD across source-domain scores within seed $s$, as in Equation~\eqref{eq:robustness}. Define
\begin{align}
I(c)&=\frac{1}{S}\sum_{s=1}^{S}I_s(c), &
R(c)&=\frac{1}{S}\sum_{s=1}^{S}r_s(c), \\
g(c)&=\max_{c'\in\mathcal C}I(c')-I(c), &
U(c)&=\sqrt{\frac{1}{S-1}\sum_{s=1}^{S}\bigl(r_s(c)-R(c)\bigr)^2}.
\label{eq:exploratory-scores}
\end{align}
The IID regret $g$ measures a candidate's source-IID loss relative to the best candidate; $U$ is the \emph{sample} SD of its robustness score across source-training seeds. Thus $U$ and the within-seed domain SD have different purposes and different denominators. IID and domain accuracies use the $[0,1]$ scale, and all penalties retain those accuracy units without additional standardization or rescaling.

Table~\ref{tab:rule-family} gives every rule in its declared order. Here $\mathcal C_k$ is the IID top-$k$ shortlist. Trust thresholds 0.005 and 0.010 correspond to 0.5 and 1.0 pp, and feasibility comparisons use numerical tolerance $10^{-12}$. The penalized family has the common form
\begin{equation}
\hat c_{\lambda,\gamma}\in\argmax_{c\in\mathcal C_{16}}
\{R(c)-\lambda g(c)-\gamma U(c)\},
\quad(\lambda,\gamma)\in\{(1,0),(2,0),(1,1)\}.
\end{equation}

\begin{table}[htbp]
\centering\small
\caption{All ten exploratory rules, in declared order. Identifiers match the archived analysis. All alternatives use the same seed-averaged LODO mean-minus-population-SD score $R$.}
\label{tab:rule-family}
\begin{tabular}{@{}lll@{}}
\toprule
Rule identifier & Eligible candidates & Maximized score \\
\midrule
\texttt{iid\_pooled} & $\mathcal C$ (all 64) & $I(c)$ \\
\texttt{lodo\_top2} & $\mathcal C_2$ & $R(c)$ \\
\texttt{lodo\_top4} & $\mathcal C_4$ & $R(c)$ \\
\texttt{lodo\_top8} & $\mathcal C_8$ & $R(c)$ \\
\texttt{lodo\_top16} & $\mathcal C_{16}$ & $R(c)$ \\
\texttt{trust\_005} & $\{c\in\mathcal C_{16}:g(c)\leq0.005\}$ & $R(c)$ \\
\texttt{trust\_010} & $\{c\in\mathcal C_{16}:g(c)\leq0.010\}$ & $R(c)$ \\
\texttt{anchor\_lambda1} & $\mathcal C_{16}$ & $R(c)-g(c)$ \\
\texttt{anchor\_lambda2} & $\mathcal C_{16}$ & $R(c)-2g(c)$ \\
\texttt{anchor\_stable} & $\mathcal C_{16}$ & $R(c)-g(c)-U(c)$ \\
\bottomrule
\end{tabular}
\end{table}

Stage C ranks IID scores in descending order, with lower configuration index breaking ties. Each selector then prefers higher selector score, higher IID score, lower IID rank, and finally lower configuration index. This differs formally from the frozen implementation's center-distance tie-breaker. Nevertheless, Stage C's top-16 variant reproduces all 14 frozen confirmation choices and target effects. Its four-dataset result also includes the two development datasets, explaining the different aggregate effect.

\subsection{All-rule outcomes and descriptive uncertainty}

Table~\ref{tab:stage-c} includes all ten outcomes. Effects first average target-training seeds within each outer task and domains within each dataset, then weight the four dataset means equally. The intervals use the paper-analysis run with 50,000 domain-within-dataset bootstrap resamples and base seed 20260925. Each rule's random stream is initialized by XORing that base seed with the big-endian integer encoded by the first four SHA-256 bytes of \texttt{selector:<rule\_identifier>}, modulo $2^{32}-1$. Interval endpoints are empirical 2.5th and 97.5th percentiles. These are descriptive exploratory intervals, not multiplicity-adjusted confirmation tests. The earlier frozen result retains its separate 20,000-resample analysis.

\begin{table}[htbp]
\centering\small
\caption{Complete exploratory comparison across four datasets and 22 tasks. Effects and intervals are pp. W/T/L counts task-level wins, ties, and losses relative to IID; the final column counts datasets with nonnegative mean effects. Rules follow Table~\ref{tab:rule-family}'s order.}
\label{tab:stage-c}
\begin{tabularx}{\textwidth}{@{}Yrrrr@{}}
\toprule
Rule & Effect & 95\% interval & W/T/L & Nonnegative \\
\midrule
Pooled IID & $0.000$ & --- & 0/22/0 & 4/4 \\
LODO within IID top 2 & $-0.236$ & $[-0.791,+0.339]$ & 5/11/6 & 3/4 \\
LODO within IID top 4 & $-0.205$ & $[-0.822,+0.447]$ & 7/6/9 & 3/4 \\
LODO within IID top 8 & $-0.371$ & $[-1.005,+0.280]$ & 8/4/10 & 2/4 \\
LODO within IID top 16 & $+0.086$ & $[-0.396,+0.593]$ & 9/0/13 & 2/4 \\
Trust region, 0.5 pp & $-0.465$ & $[-1.051,+0.118]$ & 6/5/11 & 1/4 \\
Trust region, 1.0 pp & $-0.041$ & $[-0.515,+0.481]$ & 7/1/14 & 2/4 \\
IID-regret penalty, $\lambda=1$ & $+0.062$ & $[-0.355,+0.543]$ & 7/5/10 & 3/4 \\
IID-regret penalty, $\lambda=2$ & $-0.276$ & $[-0.903,+0.401]$ & 7/5/10 & 3/4 \\
Regret and seed penalty & $-0.169$ & $[-0.744,+0.424]$ & 5/6/11 & 2/4 \\
\bottomrule
\end{tabularx}
\end{table}

\FloatBarrier
\Needspace{16\baselineskip}
\subsection{Cross-dataset rule choice}

In each retrospective leave-one-dataset-out fold, IID is always eligible. An alternative is eligible if its mean effect is nonnegative on at least two of the three training datasets and its equally weighted training-dataset mean is nonnegative. Among eligible rules, selection maximizes that mean; ties favor the best worst-dataset mean, then the fewest task-level overrides of IID, then the earliest declared rule in Table~\ref{tab:rule-family}. Only the chosen rule's effect on the held-out dataset enters the fold result.

This procedure selects top 2 when holding out VLCS, the penalty-1 rule when holding out Office-Home or TerraIncognita, and top 16 when holding out SVIRO. The held-out effects in Figure~\ref{fig:cross-dataset} average to $-0.425$ pp. Because all four datasets had already been examined during the overall investigation, this analysis characterizes retrospective sensitivity to dataset choice. It does not create a new independently confirmed selector.

\FloatBarrier
\bibliography{combined_references}
\end{document}